\documentclass{article} % For LaTeX2e
\usepackage{iclr2024_conference,times}

\iclrfinalcopy

\usepackage{amsmath,amsfonts,bm}

\def\eqref#1{equation~\ref{#1}}
\def\1{\bm{1}}

\DeclareMathAlphabet{\mathsfit}{\encodingdefault}{\sfdefault}{m}{sl}
\SetMathAlphabet{\mathsfit}{bold}{\encodingdefault}{\sfdefault}{bx}{n}

\usepackage{hyperref}

\usepackage[utf8]{inputenc} % allow utf-8 input
\usepackage[T1]{fontenc}    % use 8-bit T1 fonts
\usepackage{amsfonts}       % blackboard math symbols
\usepackage{url}
\usepackage{graphicx}
\usepackage[ethiop,main=english]{babel}

\usepackage{tipa}
\usepackage{multicol}
\usepackage{subfigure}
\usepackage{enumitem}

\title{Semantically Corrected Amharic\\Automatic Speech Recognition}

\author{Samuael Adnew\\
African Institute for Mathematical Sciences (AIMS)\\
\texttt{sadnew@aimsammi.org} \\
\And
Paul Pu Liang\\
Machine Learning Department\\
Carnegie Mellon University\\
\texttt{pliang@cs.cmu.edu}
}

\begin{document}

\maketitle
% \fontspec{Abyssinica SIL}
\begin{abstract}

Automatic Speech Recognition (ASR) can play a crucial role in enhancing the accessibility of spoken languages worldwide. In this paper, we build a set of ASR tools for Amharic, a language spoken by more than 50 million people primarily in eastern Africa. Amharic is written in the Ge'ez script, a sequence of graphemes with spacings denoting word boundaries. This makes computational processing of Amharic challenging since the location of spacings can significantly impact the meaning of formed sentences. We find that existing benchmarks for Amharic ASR do not account for these spacings and only measure individual grapheme error rates, leading to significantly inflated measurements of in-the-wild performance. In this paper, we first release corrected transcriptions of existing Amharic ASR test datasets, enabling the community to accurately evaluate progress. Furthermore, we introduce a post-processing approach using a transformer encoder-decoder architecture to organize raw ASR outputs into a grammatically complete and semantically meaningful Amharic sentence. Through experiments on the corrected test dataset, our model enhances the semantic correctness of Amharic speech recognition systems, achieving a Character Error Rate (CER) of 5.5\% and a Word Error Rate (WER) of 23.3\%.
\end{abstract}

\section{Introduction}

% Amharic and Tigre, both belonging to the Semitic language family, are primarily spoken in Ethiopia and Eritrea, respectively. While both languages employ the Ethiopic script, Tigre incorporates additional characters, albeit rarely used, derived from the Ethiopic syllabary historically utilized in Geez \citep{Dikubab_2022}.
Amharic, the official language of the Federal Democratic Republic of Ethiopia (FDRE) ~\citep{inproceedings}, is the second-largest Semitic language following Arabic and is one of the largest phonetic languages spoken in eastern Africa. The Amharic script comprises roughly 275 characters/graphemes, formed through the amalgamation of 34 consonants and 7 vowels arranged as consonant-vowel (CV) pairs. Each syllable is depicted by a singular character image (grapheme), and sentences are constructed by sequencing these characters in appropriate groupings, separated by spaces.

% Amharic is considered an agglutinative language due to its characteristic way of forming words through the addition of suffixes, prefixes, and infixes to a root or stem. These affixes carry specific meanings, such as tense, aspect, mood, gender, number, and case. Multiple morphemes can be attached to a root word to convey precise grammatical information, resulting in complex word forms. This agglutinative nature allows for a rich and nuanced expression of ideas while maintaining a relatively straightforward and systematic structure.

However, despite this significant demographic presence, there are no successful ASR models or training data available online for Amharic. One issue is the lack of abundant speech data and corresponding transcribed text to train ASR systems~\citep{interspeech}. Alongside limited data availability, the presence of Out-Of-Vocabulary (OOV) words presents a notable challenge for morphologically rich languages like Amharic~\citep{acoustic-to-word-modeling}. Therefore, prior work has attempted to use syllable-based speech recognition systems to alleviate these issues~\citep{syllable_based,first_asr}. 

Nevertheless, ASR for Amharic is fundamentally challenging because sub-word-based acoustic data extraction causes semantic and structural issues. The subword sequence output of speech recognition models in Amharic can be influenced by various factors such as variations in utterances, contextual intricacies, background noise, dialectal differences, and other relevant factors that affect the semantic correctness of the output sentences. For example, the output sequence from the ASR model, "\selectlanguage{ethiop}ha garA ^cen sa lAme na\selectlanguage{english}", contains only two meaningful words, "\selectlanguage{ethiop}garA\selectlanguage{english}" and "\selectlanguage{ethiop}lAme\selectlanguage{english}". However, these words do not align with the intended meaning of the sentence.  \selectlanguage{ethiop}"garA"\selectlanguage{english}  means "Tame" and "\selectlanguage{ethiop}lAme\selectlanguage{english}" means "Cow", when it should be "\selectlanguage{ethiop}hagarA^cen salAme nawe\selectlanguage{english}" by rectifying the spacing and adding the missing character "\selectlanguage{ethiop}we\selectlanguage{english}" at the end of the sequence which has the intended meaning: "Our country is peaceful.".

To address these issues, this paper proposes a method to semantically group Amharic sub-words into phrases and sentences. This is achieved by training an encoder-decoder transformer-based Sequence-to-Sequence (Seq2Seq) model~\citep{vaswani2017attention} to translate the output sub-word sequence of an ASR model into a complete and meaningful Amharic sentence. The translation process applies modifications such as adjusting the spacing between sub-words, inserting or removing sub-words, and changing sub-words to create a more grammatically correct and semantically coherent sentence.

\begin{figure}
    % \centering
    \includegraphics[angle=270, width=1.0\textwidth]{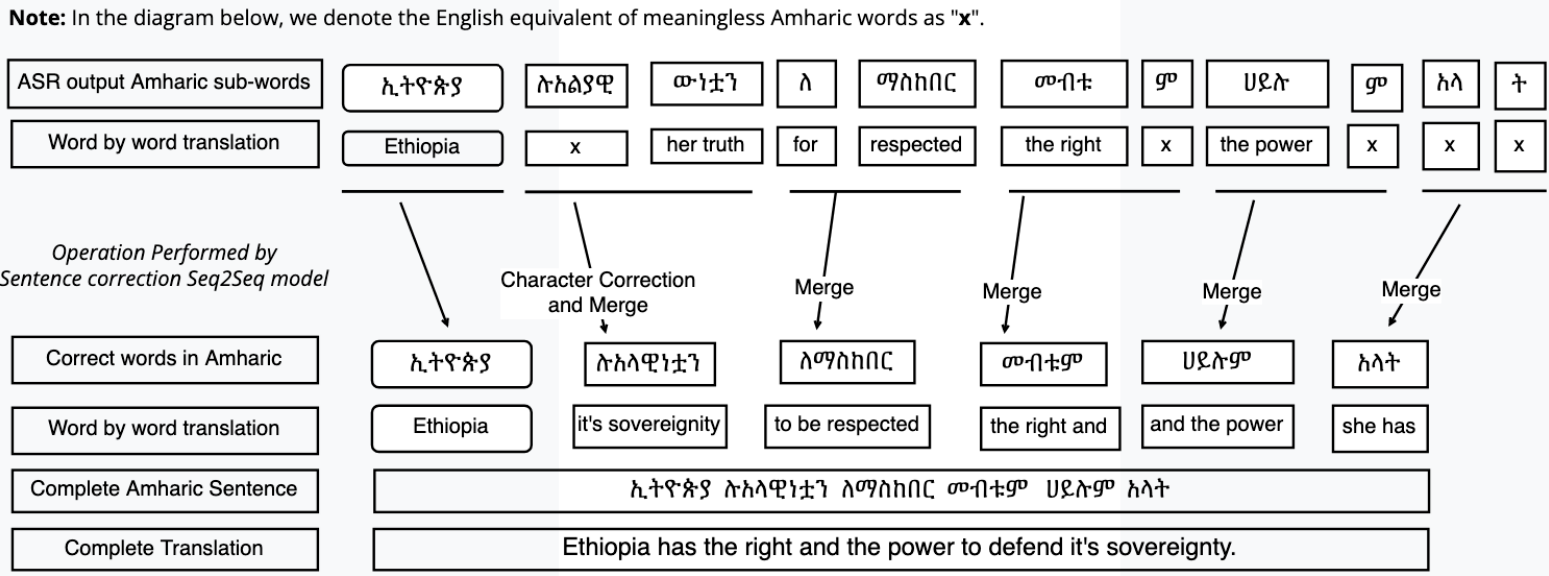}
    \caption{
    This figure illustrates the process facilitated by the Seq2Seq sentence correction model, demonstrating the restructuring of nonsensical sub-words into coherent and meaningful Amharic word sequences, culminating in the formation of a coherent sentence. Through this formal analysis, witness the progression from disorder to comprehension, ultimately leading to the generation of a refined and intelligible final output.
    }
    \label{fig:phoneme-based}
\end{figure}

We conducted a range of experiments across three base models for ASR: SpeechBrain's ASR fine-tuned for Amharic ~\citep{SB2021}, and both character and phoneme-level ASR models we created by finetuning versions of Wav2Vec2.0 models, which we term SpeechBrainASR, PhonemeAWav2Vec, and CharacterAWav2Vec respectively. On top of these base ASR models, we train Seq2Seq correction models mapping programmatically generated erroneous Amharic sentences, mimicking the errors introduced by base ASR models, to semantically correct Amharic sentences. Our resulting method achieves a Character Error Rate (CER) of 5.5\% and a Word Error Rate (WER) of 23.3\%, outperforming prior baselines both on semantic correctness and character-error-rate reduction. We release our code and models on Github: 
\url{https://github.com/samuael/postprocessed_geez_asr}.

\section{Related Work}

To date, numerous studies on end-to-end speech recognition systems have been conducted across different languages and corpora. In this section, we will review end-to-end ASR studies focusing on the sub-word level and the data construction methods employed to enhance and achieve semantically corrected speech recognition.

\paragraph{Transformer-based Language Models for ASR} While initial work had investigated n-gram~\citep{bassil2012asr} and Recurrent neural network models~\citep{guo2019spelling} for ASR, it has become increasingly popular to use sequence-to-sequence attention-based models, particularly the Transformer~\citep{vaswani2017attention}. These methods have been applied broadly to English~\citep{hrinchuk2019correction} ~\citep{automatic_gec_english}, Mandarin Chinese~\citep{syllable_based_chinese}~\citep{koo2024practical}, Arabic~\citep{kwon2023english}, Bangla~\citep{bijoy2022dpcspell}, and more modalities in text, speech, audio, video, and more~\citep{liang2022highmmt}.

\paragraph{Post-correction ASR} A vast body of previous research has focused on refining the output of ASR systems~\citep{ERRATTAHI201832}. Particularly relevant to our study,~\citep{guo2019spelling} suggests training a spelling correction model utilizing Recurrent Neural Networks (RNN) with attention mechanisms~\citep{bahdanau2016neural} to rectify errors produced by the Listen, Attend, and Spell (LAS) model. In contrast to this approach, our model adopts a Transformer architecture and eliminates the need for extra speech-to-text data generation for training. There are also approaches designed with language- or task-specific heuristics such as detecting and correcting wrong query words~\citep{sarma-palmer-2004-context}, using copy mechanisms~\citep{copy_mechanism}, performing grammatical error correction~\citep{zhou2020improving}, and tackling erroneous acoustic outputs~\citep{hrinchuk2019correction}.

\paragraph{ASR for Amharic and other low-resource languages}~\citep{phoneme} introduced an innovative approach to Amharic ASR, focusing on overcoming OOV challenges by employing a hybrid connectionist temporal classification with attention in an end-to-end architecture and phoneme-based sub-word units. ~\citep{applying_wav2vec_for_asr} employed the pre-trained wav2vec2.0 model to address low-resource ASR challenges, incorporating an extra projection layer and fine-tuning the ASR model with CTC loss. Finally, ~\citep{SB2021} combine a pre-trained wav2vec 2.0 model with two additional fine-tuned Deep Neural Network (DNN) layers followed by CTC greedy decoder for Amharic ASR, and~\citep{wang2021investigation} employed phoneme-based sub-words derived from Byte-Pair-Encoding (BPE) as modelling units for end-to-end speech recognition.

The literature above emphasizes three key points. Firstly, it discusses the effectiveness of utilizing separately trained transformer-based Seq2Seq models on extensive datasets, as opposed to limited data, to enhance the ASR model's performance and mitigate overfitting issues during training. Secondly, it highlights the use of sub-words, including characters and phonemes, as an effective approach to address Out-of-Vocabulary (OOV) challenges. Lastly, it underscores the advantages gained from fine-tuning pre-trained models such as Wav2Vec2.0, which enhances the ASR model's ability to capture patterns compared to models trained on limited data or newly created models. Our approach leverages these key insights and builds a post-correction ASR model for the Amharic language while taking into account the unique challenges and constraints for Amharic to be successful.

\section{Methodology}

We describe our method through two parts: (1) the data and fine-tuning process to adapt ASR models to Amharic and (2) the Seq2Seq sentence correction model to semantically correct transcribed Amharic outputs.

\begin{figure}
    \centering
    \vspace{-6mm}
    \includegraphics[width=0.4\linewidth]{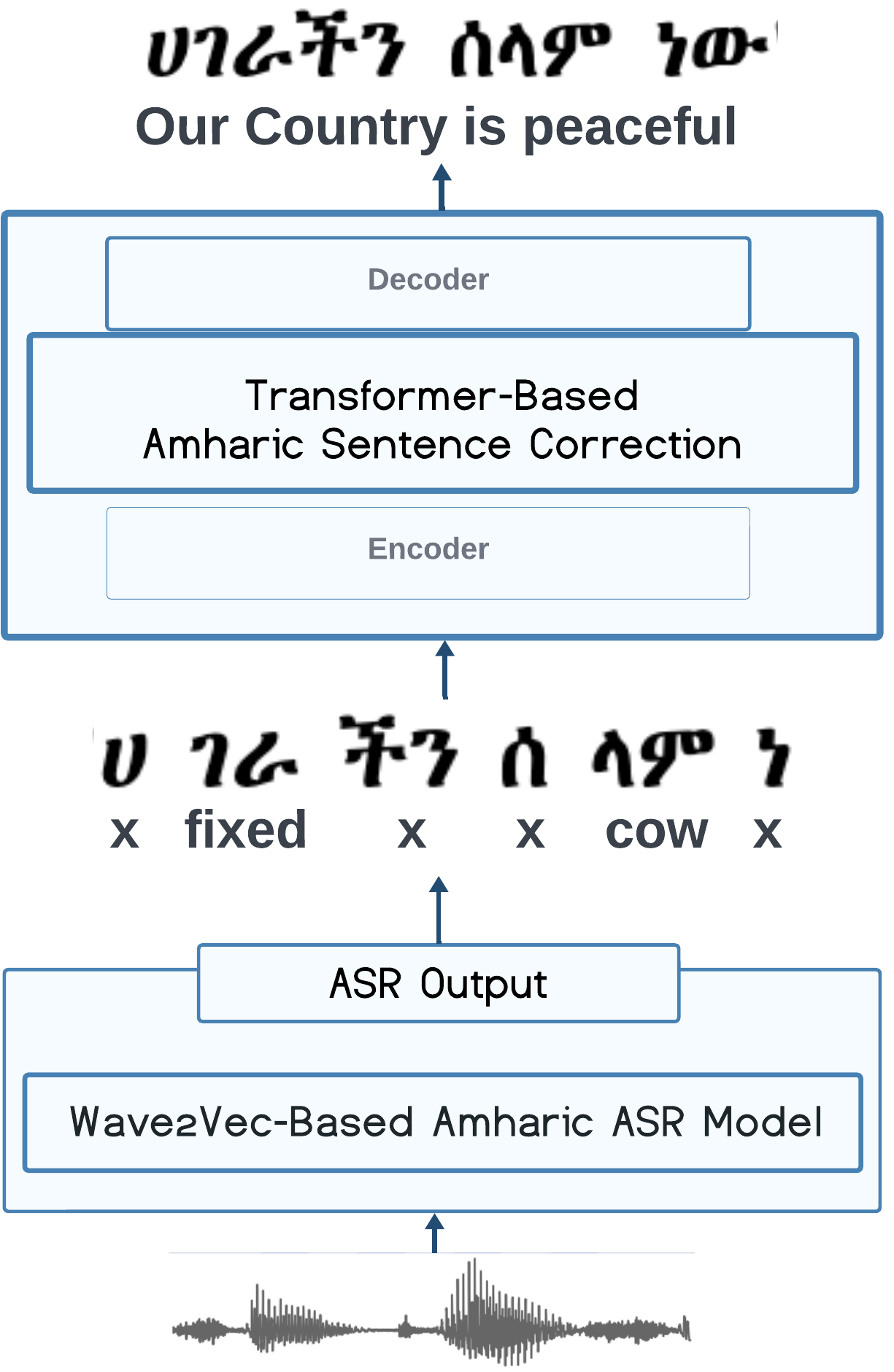}
    \caption{Our method comprises two parts: (1) the finetuned ASR model, which adapts base ASR models for Amharic, and (2) the Seq2Seq sentence correction model, designed to semantically correct transcribed Amharic outputs. In this diagram, the character 'x' is employed to denote meaningless sub-words. }
    % [TODO can be improved]} 
    \vspace{-4mm}
    \label{fig:phoneme-based}
\end{figure}

\subsection{Finetuning Wav2Vec ASR model for Amharic}

We start with a Cross-Lingual Speech Representations(XLSR) Wav2Vec2.0 model~\citep{conneau2020unsupervised} of Facebook, XLSR-Wav2Vec2, downloaded from Facebook's public repository on Huggingface. \footnote{Facebook's XLSR-Wav2Vec2 model: \href{https://huggingface.co/facebook/wav2vec2-large-xlsr-53}{https://huggingface.co/facebook/wav2vec2-large-xlsr-53}}. Originally XLSR-Wav2Vec2 is a large model pre-trained on 53 languages hours of Multilingual Librispeech(MLS), CommonVoice, and BABEL datasets, covering a duration of over 50k hour of speech, and utilizes a convolutional feature encoder and a transformer to process raw waveforms to generate fixed-size speech representations. To adapt it for Amharic, we fine-tune it with more than 100 hours, and 10875 data-points of Amharic data. This results in two models: PhonemeAWav2Vec and CharacterAWav2Vec. These models perform ASR at the phoneme and character-level, respectively, which alleviates the OOV issues typically seen in low-resource languages.

\paragraph{Data curation} Specifically, we fine-tune wav2vec for Amharic using the ALFFA (African Languages in the Field: Speech Fundamentals and Automation) Amharic reading speech database ~\citep{tachbelie2014}, featuring 10,875 labeled audio recordings totaling over 100 hours for speech recognition tasks. To enhance the model's robustness to external noise, we performed data augmentation on this dataset with white noise and real-life noise, resulting in 25,508 samples, including 14,633 additional augmented samples. This augmentation involved adding randomly sampled white noise and, at most, three additional reading samples from our training data as background noise. For testing, we performed semantic and structural corrections and utilized a separate test dataset of 359 data points provided by ALFFA to evaluate performance.

\paragraph{Training details} By developing these Wav2Vec2.0 models for ASR at the phoneme and character-level, respectively, we alleviate the OOV issues typically seen in low-resource languages. Our phoneme-level model is called PhonemeAWav2Vec, consisting used 27 base consonants, each representing a unique base sound, and 7 vowels to denote Amharic syllables for a total vocabulary size of 33 (details in Appendix~\ref{app:ethiopia_vowels}). Our character-level model is called CharacterAWav2Vec, consisting of 197 basic syllables and 33 derived syllables for a total vocabulary size of 233 (details in Appendix~\ref{app:additional_details}).

For the PhonemeAWav2Vec ASR model, we selected the XLSR Wav2Vec2ForCTC model, which is built upon the Wav2VecForCTC architecture pre-trained using speech data from multiple languages. Training was conducted using an A100 GPU for approximately 52.64 epochs, employing the Adam optimizer with a learning rate of 3e-05, epsilon of 1e-08, and batch sizes of 32 for training and 8 for evaluation. The fine-tuning training process continued for 42,000 steps with linear decay in the learning rate.\footnote{Our phoneme-level model: \href{https://huggingface.co/Samuael/asr-amharic-phoneme-based-37-6}{https://huggingface.co/Samuael/asr-amharic-phoneme-based-37-6}} \footnote{Our character-level model: \href{https://huggingface.co/Samuael/asr-amharic-phoneme-based-233}{https://huggingface.co/Samuael/asr-amharic-phoneme-based-233}}

To train the CharacterAWav2Vec model, we duplicated the PhonemeAWav2Vec model, adjusted the output a vocabulary size of 233 to represent all uniquely sounded Amharic syllables, and then further fine-tuned the modified model. Training also utilized an A100 GPU, lasting approximately 12.78 epochs, and employed the Adam optimizer with an initial learning rate of 4e-06 and epsilon set to 1e-08. Similar batch sizes were used for both training and evaluation, with the training proceeding for a maximum of 2400 steps and linear decay in the learning rate.

\subsection{Seq2Seq sentence correction model dataset and training}
The third and final component of our model addresses the challenge of post-correcting ASR outputs, which often suffer from incorrect spacing between sub-words or characters, as well as missing or mis-detected sub-words. These errors can lead to semantic inaccuracies and make the output difficult to comprehend. Our solution involves grouping sub-words to form meaningful words, thus resolving these issues. This post-correction model takes as input sequences from speech recognition outputs and returns properly organized and corrected sentences that are readable. We achieve this through a sequence-to-sequence transformer model ~\citep{vaswani2017attention}, employing an encoder-decoder structure. The encoder processes the incoming character sequence, while the decoder 'translates' it into a corrected sequence, ensuring improved coherence and accuracy.

\paragraph{Data curation} We trained the correction model using a text-only corpus gathered from the Holy Bible, fiction, and various news outlets~\citep{amharictextcorpus, unknown}. Programmatically, we generated flawed text to simulate the errors made by ASR models, resulting in a total of 90,927 sentence and correction pairs. To prepare the training data, we conducted normalization, removing unnecessary characters and symbols. Additionally, we substituted structurally distinct characters with similar phonetic sounds, aligning them with our carefully curated list of uniquely voiced characters. These uniquely curated voiced characters were selected due to Amharic's several groups of characters with different graphical representations but similar sounds. For example, characters such as '\selectlanguage{ethiop}ha\selectlanguage{english}', '\selectlanguage{ethiop}hA\selectlanguage{english}', '\selectlanguage{ethiop}.ha\selectlanguage{english}', and '\selectlanguage{ethiop}.hA\selectlanguage{english}' all produce the 'ha' sound and are interchangeable based on certain conventions. For instance, when writing 'Country,' the first '\selectlanguage{ethiop}ha\selectlanguage{english}' character is typically used, constructing the word as '\selectlanguage{ethiop}hagar\selectlanguage{english}'. Conversely, when writing 'Lake,' either '\selectlanguage{ethiop}.ha\selectlanguage{english}' or '\selectlanguage{ethiop}.hA\selectlanguage{english}' is often utilized, as these characters resemble and represent a ship. In our experiment, we normalized the dataset to utilize the most commonly used character to represent other characters with the same sound in syllables/characters.

\paragraph{Training details} We have designed these seq2seq correction models with token sizes of 233 and 15K tokens to compare the performance of Seq2Seq models with different token sizes in correcting malformed sentences. Specifically, we have named the Seq2Seq transformer model with a token size of 233 as Seq2Seq233. Additionally, the Seq2Seq transformer model, featuring a token size of 15,100, includes both smaller and larger variants, referred to as \textbf{Seq2SeqT5} and \textbf{Seq2SeqBT5}, respectively.

For \textbf{Seq2Seq233}, we used a d-model of 256, 8 layers, and 12 attention heads, each having key, query, and value sizes of 32. We set tokens in batch to 10000 and batches per step to 25000. The initial learning rate is 1.75e-07, betas are set to (0.9, 0.98), epsilon is 1e-9, and dropout is 0.1, with Adam optimizer~\citep{kingma2017adam} and loss function of mean label-smoothed cross-entropy loss. The training spans a total of 100 epochs, aiming to generate semantically correct Amharic sentences from sequences of selected Ethiopic characters.

For \textbf{Seq2SeqT5}, we employed the Text-to-Text Transfer Transformer (T5) model ~\citep{raffel2023exploring} with the default configuration for conditional generation, T5ForConditionalGeneration, in our Seq2Seq transformer model with a token size of 15K. This model architecture is based on a transformer featuring 6 layers with a model dimension of 512 and 8 attention heads. The decoder comprises 8 layers, with a feed-forward dimension of 2048, and key and value dimensions of 64. ReLU activation functions are utilized for dense layers and feed-forward projections. The vocabulary size is set to 15100, including special tokens. The training was conducted for 1000 steps with a batch size of 128 on a Google Colab A100 GPU, achieving a validation loss of 0.0864.

Regarding \textbf{Seq2SeqBT5}, we retained the same model architecture and token size with Seq2SeqT5, albeit with a model consisting of 12 layers and a model dimension (d-model) of 768. Training extended over 29,311 steps with a batch size of 64, conducted on a Google Colab L4 GPU, ultimately yielding a validation loss of 0.801.
\footnote{find our Seq2SeqT5 models here: \href{https://huggingface.co/Samuael/geez_t5-15k}{https://huggingface.co/Samuael/geez\_t5-15k}
}
\footnote{
find our Seq2SeqBT5 model here:
\href{https://huggingface.co/Samuael/geez_t5-big-15k}{https://huggingface.co/Samuael/geez\_t5-big-15k}
}

\section{Results}

In our evaluation, we utilized the CER and WER metrics to assess the error rates of the output sentences. This assessment was conducted using the ALFFA Amharic test dataset. We manually rectified the testing dataset sourced from ALFFA Amharic to establish a semantically corrected evaluation dataset. Our efforts were prompted by the discovery of numerous errors within the original data, including the addition of spacing to align utterances on the input audio with the sub-words or syllables on the transcriptions, resulting in incorrect sentence structure, as well as missing or erroneous syllables, which made prior evaluations on the test set somewhat unreliable.
\footnote{\href{https://github.com/samuael/postprocessed_geez_asr/blob/main/SemanticallyCorrectedTestDataset/}{https://github.com/samuael/postprocessed\_geez\_asr/blob/main/SemanticallyCorrectedTestDataset/}}

\begin{table}[t]
\centering
\def\arraystretch{1.2}%
\begin{tabular}{c | c c}
 \hline \hline
 Model Name & CER & WER \\
 \hline
 \multicolumn{3}{c}{\textbf{vanilla ASR models, tested on incorrect data}} \\ 
 \hline
 SpeechBrainASR & 6.6\% & 24.9\%\\ 
 PhonemeAWav2Vec & 7.7\% & 28.8\% \\
 CharacterAWav2Vec & 7.4\% & 27.1\% \\
 \hline
 \multicolumn{3}{c}{\textbf{vanilla ASR models, tested on corrected data}} \\ 
 \hline
 SpeechBrainASR & 14.1\% & 96.6\%\\ 
 PhonemeAWav2Vec & 14.3\% & 99.9\% \\
 CharacterAWav2Vec & 14.0\% & 98.6\% \\
 \hline
 \multicolumn{3}{c}{\textbf{our best model, tested on corrected data}} \\ 
 \hline
 SpeechBrainASR + ours &    \textbf{5.9}\% & \textbf{24.9}\%  \\ 
 PhonemeAWav2Vec + ours &   \textbf{5.5}\% & \textbf{23.3}\%  \\
 CharacterAWav2Vec + ours & \textbf{5.6}\%  & \textbf{23.4}\% \\
 \hline \hline
\end{tabular}
\caption{The first 3 rows show vanilla ASR models evaluated on original incorrect datasets, which severely overestimate actual model capabilities. The next 3 rows show a more realistic measure of model performance on corrected datasets - we find that they struggle with very high error rates. After applying our post-correction methods (final 3 rows), we achieved significantly improved performance on the semantically corrected test datasets, with new state-of-the-art ASR CER\% and WER\% for Amharic ASR. }
% [TODO fill table in]}
\label{app:table_first}
\end{table}

\begin{table}[t]
\centering
\def\arraystretch{1.2}%
\begin{tabular}{c | c c}
 \hline \hline
 Model Name & CER & WER \\
 \hline
 \multicolumn{3}{c}{\textbf{with Seq2Seq233 correction, tested on corrected data}} \\ 
 \hline
 SpeechBrainASR + Seq2Seq233 & 7.4\% & 28.7\%  \\ 
 PhonemeAWav2Vec + Seq2Seq233 & 7.0\% & 27.3\%  \\
 CharacterAWav2Vec + Seq2Seq233 & 7.1\%  & 27.8\% \\
 \hline
 \multicolumn{3}{c}{\textbf{with Seq2SeqT5 correction, tested on corrected data}} \\ 
 \hline
 SpeechBrainASR + Seq2SeqT5 & 6.9\% & 26.2\%  \\ 
 PhonemeAWav2Vec + Seq2SeqT5 & 6.7\% & 25.4\%  \\
 CharacterAWav2Vec + Seq2SeqT5 & 6.5\%  & 24.8\% \\
 \hline \hline
\multicolumn{3}{c}{\textbf{with Seq2SeqBT5 correction, tested on corrected data}} \\ 
 \hline
 SpeechBrainASR + Seq2SeqBT5 &    5.9\% & 24.9\%  \\ 
 PhonemeAWav2Vec + Seq2SeqBT5 &   5.5\% & 23.3\%  \\
 CharacterAWav2Vec + Seq2SeqBT5 & 5.6\%  & 23.4\% \\
 \hline \hline
\end{tabular}
\caption{Ablation studies: We evaluate the impact of various base ASR models (SpeechBrainASR, PhonemeAWav2Vec, CharacterAWav2Vec) and different Transformer-based Encoder-Decoder Seq2Seq sentence correction methods, achieving a 5.5\% CER and 23.3\% WER, on overall performance. 
% [TODO fill this in]
}
\label{app:table_all}
\end{table}

\paragraph{Baselines}

We compare three types of approaches:
\begin{enumerate}[noitemsep,topsep=0pt,nosep,leftmargin=*,parsep=0pt,partopsep=0pt]
    \item Vanilla ASR models tested on original data without accounting for the lack of spacing. We use a publicly available model of \textbf{SpeechBrain} ~\citep{SB2021}, an End-to-End (E2E) convolutional model for ASR, fine-tuned on a publicly available dataset for Amharic ASR\footnote{SpeechBrain Amharic ASR: \href{https://huggingface.co/speechbrain/asr-wav2vec2-dvoice-amharic}{https://huggingface.co/speechbrain/asr-wav2vec2-dvoice-amharic}}. The SpeechBrain architecture processes mel-filter bank features derived from the wav2vec2.0~\citep{baevski2020wav2vec} acoustic model and uses Connectionist Temporal Classification (CTC) to align speech and language sequences.The results are presented in the first three rows of Table~\ref{app:table_first}.
    As explained above, we utilize two additional vanilla ASR models, namely PhonemeAWav2Vec and CharacterAWav2Vec, and apply a similar approach in this and the following two processes.
    
    \item Vanilla ASR models testing using corrected data. We take the same SpeechBrain Amharic ASR model as described above and evaluate it on the semantically correct and properly spaced testing dataset, to give an estimate of true performance on corrected data.
    
    \item After obtaining the output of ASR models, we proceeded to apply character-based semantic correction using the Seq2Seq233 model, as well as subword-based correction using the Seq2SeqT5 and Seq2SeqBT5 models, as part of our approach.

     % TODO what 3 phases mean? also u explain Seq2Seq233 and Seq2SeqT5 but the table also includes Seq2SeqBT5 what is that?
    % TODO we should separate table 1 into 2 - first table only has vanilla ASR models rows and the BEST of our method. second table can ablate Seq2Seq233 vs Seq2SeqT5 vs  Seq2SeqBT5 because its all variants of our method.
\end{enumerate}

\subsection{Main results}

We show the main comparisons in Table~\ref{app:table_first}. Our primary findings reveal that our method achieves a Character Error Rate of 5.5\% and a Word Error Rate of 23.3\%, representing significant improvements over the publicly available baseline model of \textbf{SpeechBrain} for Amharic. Our model offers the following enhancements:

\begin{enumerate}[noitemsep,topsep=0pt,nosep,leftmargin=*,parsep=0pt,partopsep=0pt]
\item Improved semantic correctness and readability of speech recognition output are achieved by rectifying erroneously grouped or spaced sub-words, ensuring proper spacing to form meaningful words and sequences that convey the intended meaning effectively.

\item Enhanced sub-word recognition resulted in an 8.49\% reduction in Character Error Rate on the semantically corrected testing dataset compared to the ASR outputs.
\end{enumerate}

\paragraph{Qualitative results} Below, we show sample outputs of speech recognition both before and after Seq2Seq correction: 

Before Correction: "\selectlanguage{ethiop}ya 'al ba^ser manege st ka zih ba_huAlA ya'ErtrA lu'a lAwi nat yami.sArare nA ya bAlasl.tAnAt wn  s b'enA kami nakA tagbAr maqo .tab 'endami nor bat gal.sawAle\selectlanguage{english}" \\

After Correction: "\selectlanguage{ethiop}ya'alba^ser manegest kazih ba_huAlA ya'ErtrA lu'alAwinat yami.sArarenA yabAlasl.tAnAtn  sb'enA kaminakA tagbAr maqo.tab 'endaminorbat gal.sawAle\selectlanguage{english}"\\
% የአልበሽር መንግስት ከዚህ በኋላ የኤርትራ ሉአላዊነት የሚጻረርና የባለስልጣናትን ውስብእና ከሚነካ ተግባር መቆጠብ እንደሚኖርበት ገልጸዋል

Meaning: "Al-Bashir's government has stated that from now on, it will refrain from actions that go against the sovereignty of Eritrea and affect the reputation of the officials."

\begin{center}
\line(1,0){70}
\end{center}
% \newline
Before Correction: "\selectlanguage{ethiop}
mab rAt hay l ba wsAnEw ta.taqAmi sAyhn 'enda mAyqar ka zEnA mn^Co^cA^cn  lamaradA^cn ^clanAl\selectlanguage{english}"\\

% መብ ራት ሀይ ል በ ውሳኔው ተጠቃሚ ሳይህን እንደ ማይቀር ከ ዜና ምንጮቻችን ለ መረዳ ችን ለናል

 After Correction: "
 \selectlanguage{ethiop}mabrAt hayl bawsAnEw ta.taqAmi sAyhon 'endamAyqar kazEnA mn^Co^cA^cn  lamaradAt ^clanAl\selectlanguage{english}"\\
 
 % መብራት ሀይል በውሳኔው ተጠቃሚ ሳይሆን እንደማይቀር ከ ዜና ምንጮቻችን ለመረዳችን ችለናል
 
 Meaning: "We have been able to understand from our news sources that the electric utility service bureau will not be the beneficiary of the decision"
 % \newline
 \begin{center}
\line(1,0){70}
\end{center}

\subsection{Ablation studies}

To demonstrate the individual contributions of each component to the overall improvements, we perform extensive ablations on 2 parts: the base ASR model and the various Seq2Seq architectures for semantic correction. 
As depicted in Table~\ref{app:table_all}, the individual speech extraction models — namely, \textit{SpeechBrainASR}, \textit{PhonemeAWav2Vec}, and \textit{CharacterAWav2Vec} — show disparate error rate outcomes both pre- and post-application of the \textit{Seq2Seq} sentence correction. Notably, the Seq2SeqBT5 model demonstrates superior performance across all three ASR models, with the PhonemeAWav2Vec output corrected by the Seq2SeqBT5 yielding the most favorable result.
% [TODO this section needs work]

\paragraph{Ablations on base ASR model} 

The output of PhonemeAWav2Vec, evaluated on the corrected test dataset, resulted in a CER of 14.3\% and WER of 99.9\%. This outcome suggests the presence of character errors and inadequate spacing between character groups, resulting in very few correctly recognized words compared to what a structurally correct Amharic sentence should contain. A similar pattern is observed in the other speech extraction models, CharacterAWav2Vec and SpeechBrainASR, indicating that these speech recognition models struggle with incorrectly detected sub-words/characters and improper spacing, thus leading to high WER scores.

\paragraph{Ablations on Seq2Seq model architecture} 
% TODO... can we split this section into 2 parts..  copy stuff below into the right part

A consistent trend of improved accuracy is evident across various components of speech recognition models following the application of Seq2Seq models. Particularly noteworthy is the proficiency of the Seq2SeqBT5 model in managing proper spacing between sequence characters, resulting in a significant reduction in both error-rate metrics.
For instance, when applying the Seq2SeqBT5 model to the output of PhonemeAWav2Vec ASR, we achieved a CER of 5.5\% and a WER of 23.3\%. These results represent reductions in CER by 8.5\% and WER by 76.6\% compared to the lowest error rates recorded by the vanilla ASR models measured on the corrected test dataset.

% It becomes necessary to apply the Seq2Seq transformer model to correct ('translate') the poorly structured character sequence into its proper form and make updates on the characters to reduce the CER, consequently reducing the WER.
%The findings presented above demonstrate a notable decrease in error rates, underscoring the effectiveness of Seq2Seq in accurately identifying and rectifying sub-word issues within the input sequence. All of the Seq2Seq models performed well, with the Seq2SeqBT5 model achieving the most significant reduction in error rates. This further highlights its superior performance in addressing sub-word issues within the input sequence.
% [TODO which type of seq2seq model? or all of them?]
% This encompassed tasks performed by the Seq2Seq encoder-decoder transformer includes adding, replacing, or removing join elements to enhance sentence structure. 

% [Moved to future work section]

% [TODO i don't understand this paragraph .. what are you trying to say]

\section{Conclusion}

Overall, we achieved a CER of 5.5\% and a WER of 23.3\%, a notable improvement of 8.49\% in CER compared to the outputs of ASR models on the corrected testing dataset. Our study highlights the efficacy of integrating a pre-trained Seq2Seq model for sentence correction on sub-word-based speech recognition models, yielding benefits in generalizing to OOV words while generating semantically correct outputs. Furthermore, the modular structure we have adopted not only allows for independent training of model components but also provides the flexibility to extend the training of the existing Seq2Seq model to encompass other languages utilizing the Ethiopic script.

\paragraph{Future Work} Our investigation into the semantically corrected ASR model for the Amharic language is ongoing, and we plan to conduct additional experiments to enhance accuracy. One avenue for improvement includes incorporating punctuation marks to bolster the robustness of our models. Moreover, the adoption of the state-of-the-art Grammatical Error Correction (GEC) approach~\citep{Bryant_2023} has shown promising results in sentence correction and should/can be explored for further performance enhancements.
Additionally, With further modifications and fine-tuning of existing speech extraction models for Amharic, along with additional training of Seq2Seq models using various languages utilizing the Ethiopic script to refine ASR outputs, this research can be extended to support other languages employing the Ethiopic script spoken in Ethiopia and Eritrea.

\paragraph{Data Availability Statement}

The ALFFA dataset is accessible at the following link: \href{https://www.openslr.org/25/}{https://www.openslr.org/25/}.

\newpage

\bibliography{iclr2024_conference}
\bibliographystyle{iclr2024_conference}

\newpage

\section{Appendix A: Additional Details}

\subsection{Ethiopic Vowels}\label{app:ethiopia_vowels}

\textit{Table 3:The seven(7) vowels used in Amharic script.}
\begin{center}
\begin{tabular}{| c | c c c c c c c |} 
 \hline
 Grapheme &\selectlanguage{ethiop}'ua\selectlanguage{english}&\selectlanguage{ethiop}'u\selectlanguage{english}&\selectlanguage{ethiop}'i\selectlanguage{english}&\selectlanguage{ethiop}'A\selectlanguage{english}&\selectlanguage{ethiop}'E\selectlanguage{english}&\selectlanguage{ethiop}'e\selectlanguage{english}&\selectlanguage{ethiop}'o\selectlanguage{english}\\ [0.05ex] 
 \hline
 
 % Sound & \Əə	& u &	i&	a&	e&	ɨ&	o \\
 Sound & \textschwa	& u &	i&	a&	e&	\textbari{} &	o \\
 
\hline
\end{tabular}
\end{center}

\subsection{Used Ethiopic Consonants}\label{app:ethiopia_consonants}

Note: While the following consonants represent their respective groups, it is important to note that they do not encompass all the consonants utilized in the ASR model.

\textit{Table 4: Sixth order Amharic base characters}

\begin{center}
\begin{tabular}{|c| c c c c c c c c c c c c c c||} 
 \hline
 Grapheme &\selectlanguage{ethiop}h\selectlanguage{english}&\selectlanguage{ethiop}l\selectlanguage{english}&\selectlanguage{ethiop}m\selectlanguage{english}&\selectlanguage{ethiop}r\selectlanguage{english}&\selectlanguage{ethiop}s\selectlanguage{english}&\selectlanguage{ethiop}^s\selectlanguage{english}&\selectlanguage{ethiop}q\selectlanguage{english}&\selectlanguage{ethiop}b\selectlanguage{english}&\selectlanguage{ethiop}v\selectlanguage{english}&\selectlanguage{ethiop}t\selectlanguage{english}&\selectlanguage{ethiop}^c\selectlanguage{english}&\selectlanguage{ethiop}n\selectlanguage{english}&\selectlanguage{ethiop}~n\selectlanguage{english}&\selectlanguage{ethiop}k\selectlanguage{english} \\ [0.05ex] 
 \hline
 % Sound &h&l&m&r&ś&š&q&b&v&t&č&n&ñ&k \\
 Sound &h&l&m&r&\'{s}&\v{s}&q&b&v&t&\v{c}&n&\~{n}&k \\
 \hline
 \hline
 Grapheme &\selectlanguage{ethiop}w\selectlanguage{english}&
\selectlanguage{ethiop}z\selectlanguage{english}&
\selectlanguage{ethiop}^z\selectlanguage{english}&
\selectlanguage{ethiop}y\selectlanguage{english}&
\selectlanguage{ethiop}d\selectlanguage{english}&
\selectlanguage{ethiop}^g\selectlanguage{english}&
\selectlanguage{ethiop}g\selectlanguage{english}&
\selectlanguage{ethiop}.t\selectlanguage{english}&
\selectlanguage{ethiop}^C\selectlanguage{english}&
\selectlanguage{ethiop}.p\selectlanguage{english}&
\selectlanguage{ethiop}.s\selectlanguage{english}&
\selectlanguage{ethiop}f\selectlanguage{english}&
\selectlanguage{ethiop}p\selectlanguage{english}&
\selectlanguage{ethiop}_k\selectlanguage{english}\\ [0.05ex]
 \hline
 % Sound &w&z&ž&y&d&j&g&ṭ&ċ&pe&ṣ&f&p&kh \\
 Sound &w&z&\v{z}&y&d&j&g&\d{t}&\.{c}&pe&\d{s}&f&p&kh \\
 \hline
\end{tabular}
\end{center}

These consonants together with the 7 vowels form their own character group, with each group consisting of 7 syllable representing Consonant-Vowel (CV) combinations.

\subsection{Chart of Amharic Grapheme}\label{app:ethiopia_consonant-to-vowels}

\textit{Table 5: Chart depicting the combinations of Amharic consonants and vowels that form syllables, each represented by a distinct grapheme.}

\begin{center}
\begin{tabular}{|c| c c c c c c c|}
 \hline
  &  & &	Vowels & &	&	& \\ [0.1ex] 
Consonants & 
\textbf{\textschwa}	& u &	i &	a&	e &	\textbf{\textbari{}} &	o \\ [0.1ex] 
 \hline
 h & \selectlanguage{ethiop}ha\selectlanguage{english} & \selectlanguage{ethiop}hu\selectlanguage{english} & \selectlanguage{ethiop}hi\selectlanguage{english} & \selectlanguage{ethiop}hA\selectlanguage{english} & \selectlanguage{ethiop}hE\selectlanguage{english} & \selectlanguage{ethiop}he\selectlanguage{english} & \selectlanguage{ethiop}ho\selectlanguage{english}\\ [0.05ex]
 l & \selectlanguage{ethiop}la\selectlanguage{english} & \selectlanguage{ethiop}lu\selectlanguage{english} & \selectlanguage{ethiop}li\selectlanguage{english} & \selectlanguage{ethiop}lA\selectlanguage{english} & \selectlanguage{ethiop}lE\selectlanguage{english} & \selectlanguage{ethiop}le\selectlanguage{english} & \selectlanguage{ethiop}lo\selectlanguage{english}\\ [0.05ex]
 m & \selectlanguage{ethiop}ma\selectlanguage{english} & \selectlanguage{ethiop}mu\selectlanguage{english} & \selectlanguage{ethiop}mi\selectlanguage{english} & \selectlanguage{ethiop}mA\selectlanguage{english} & \selectlanguage{ethiop}mE\selectlanguage{english} & \selectlanguage{ethiop}me\selectlanguage{english} & \selectlanguage{ethiop}mo\selectlanguage{english}\\ [0.05ex]
 \'s & \selectlanguage{ethiop}sa\selectlanguage{english} & \selectlanguage{ethiop}su\selectlanguage{english} & \selectlanguage{ethiop}si\selectlanguage{english} & \selectlanguage{ethiop}sA\selectlanguage{english} & \selectlanguage{ethiop}sE\selectlanguage{english} & \selectlanguage{ethiop}se\selectlanguage{english} & \selectlanguage{ethiop}so\selectlanguage{english}\\ [0.05ex]
 r & \selectlanguage{ethiop}ra\selectlanguage{english} & \selectlanguage{ethiop}ru\selectlanguage{english} & \selectlanguage{ethiop}ri\selectlanguage{english} & \selectlanguage{ethiop}rA\selectlanguage{english} & \selectlanguage{ethiop}rE\selectlanguage{english} & \selectlanguage{ethiop}re\selectlanguage{english} & \selectlanguage{ethiop}ro\selectlanguage{english}\\ [0.05ex]
 \v{s} & \selectlanguage{ethiop}^sa\selectlanguage{english} & \selectlanguage{ethiop}^su\selectlanguage{english} & \selectlanguage{ethiop}^si\selectlanguage{english} & \selectlanguage{ethiop}^sA\selectlanguage{english} & \selectlanguage{ethiop}^sE\selectlanguage{english} & \selectlanguage{ethiop}^se\selectlanguage{english} & \selectlanguage{ethiop}^so\selectlanguage{english}\\ [0.05ex]
 
 . &  &  &  & . &  &  & \\ [0.05ex]
 . &  &  &  & . &  &  & \\ [0.05ex]
 . &  &  &  & . &  &  & \\ [0.05ex]
 . &  &  &  & . &  &  & \\ [0.05ex]
 f & \selectlanguage{ethiop}fa\selectlanguage{english} & \selectlanguage{ethiop}fu\selectlanguage{english} & \selectlanguage{ethiop}fi\selectlanguage{english} & \selectlanguage{ethiop}fA\selectlanguage{english} & \selectlanguage{ethiop}fE\selectlanguage{english} & \selectlanguage{ethiop}fe\selectlanguage{english} & \selectlanguage{ethiop}fo\selectlanguage{english}\\ [0.05ex]
 p & \selectlanguage{ethiop}pa\selectlanguage{english} & \selectlanguage{ethiop}pu\selectlanguage{english} & \selectlanguage{ethiop}pi\selectlanguage{english} & \selectlanguage{ethiop}pA\selectlanguage{english} & \selectlanguage{ethiop}pE\selectlanguage{english} & \selectlanguage{ethiop}pe\selectlanguage{english} & \selectlanguage{ethiop}po\selectlanguage{english}\\ [0.05ex]
 \hline
\end{tabular}
\end{center}

\subsection{Sample basic Ethiopic phoneme syllable representation}\label{app:char_to_phoneme}

Examples illustrating the representation of consonant-vowel combinations (syllables) through graphemes, with the base consonant '\selectlanguage{ethiop}l\selectlanguage{english}' (l) as the reference.

\textit{Table 6: Sample Amharic base characters for letter group '\selectlanguage{ethiop}la\selectlanguage{english}'}

\begin{center}
\begin{tabular}{|c| c  c  c  c  c  c  c |} 
 \hline
 Grapheme &\selectlanguage{ethiop}la\selectlanguage{english} & \selectlanguage{ethiop}lu\selectlanguage{english} & \selectlanguage{ethiop}li\selectlanguage{english} & \selectlanguage{ethiop}lA\selectlanguage{english} & \selectlanguage{ethiop}lE\selectlanguage{english} & \selectlanguage{ethiop}le\selectlanguage{english} & \selectlanguage{ethiop}lo\selectlanguage{english}\\ [0.05ex] 
 \hline
 Phoneme representation & \selectlanguage{ethiop}l'ua\selectlanguage{english} & \selectlanguage{ethiop}l'u\selectlanguage{english} & \selectlanguage{ethiop}l'i\selectlanguage{english} & \selectlanguage{ethiop}l'A\selectlanguage{english} & \selectlanguage{ethiop}l'E\selectlanguage{english} & \selectlanguage{ethiop}l'e\selectlanguage{english} & \selectlanguage{ethiop}l'o\selectlanguage{english}\\
 \hline
\end{tabular}
\end{center}

\newpage

\subsection{Sample Derived Ethiopic Consonants and Pronunciation}\label{app:additional_details}

Sample derived Ethiopic characters used in our vocabulary.

\textit{Table 7: Sample Derived characters}
\begin{center}
\begin{tabular}{|c| c  c  c  c |} 
 \hline
 Syllable & Consonant & vowel-1 & vowel-2 & Pronunciation\\ [0.05ex]
 \hline
\selectlanguage{ethiop}huA\selectlanguage{english} & h(\selectlanguage{ethiop}h\selectlanguage{english}) & u(\selectlanguage{ethiop}'u\selectlanguage{english}) & a(\selectlanguage{ethiop}'A\selectlanguage{english}) & h-u-a \\ [0.05ex]
\selectlanguage{ethiop}luA\selectlanguage{english} & l(\selectlanguage{ethiop}l\selectlanguage{english}) & u(\selectlanguage{ethiop}'u\selectlanguage{english}) & a(\selectlanguage{ethiop}'A\selectlanguage{english}) & l-u-a\\ [0.05ex]
\selectlanguage{ethiop}muA\selectlanguage{english} & m(\selectlanguage{ethiop}m\selectlanguage{english}) & u(\selectlanguage{ethiop}'u\selectlanguage{english}) & a(\selectlanguage{ethiop}'A\selectlanguage{english}) &  m-u-a \\ [0.05ex]
\selectlanguage{ethiop}suA\selectlanguage{english} & \'{s}(\selectlanguage{ethiop}s\selectlanguage{english}) & u(\selectlanguage{ethiop}'u\selectlanguage{english}) & a(\selectlanguage{ethiop}'A\selectlanguage{english}) &  \'{s}-u-a \\ [0.05ex]
\selectlanguage{ethiop}^suA\selectlanguage{english} & \'{s}(\selectlanguage{ethiop}^s\selectlanguage{english}) & u(\selectlanguage{ethiop}'u\selectlanguage{english}) & a(\selectlanguage{ethiop}'A\selectlanguage{english}) &  \v{s}-u-a \\ [0.05ex]
\selectlanguage{ethiop}quA\selectlanguage{english} & q(\selectlanguage{ethiop}q\selectlanguage{english}) & u(\selectlanguage{ethiop}'u\selectlanguage{english}) & a(\selectlanguage{ethiop}'A\selectlanguage{english}) &  q-u-a \\ [0.05ex]
\selectlanguage{ethiop}buA\selectlanguage{english} & b(\selectlanguage{ethiop}b\selectlanguage{english}) & u(\selectlanguage{ethiop}'u\selectlanguage{english}) & a(\selectlanguage{ethiop}'A\selectlanguage{english}) &  b-u-a \\ [0.05ex]
\selectlanguage{ethiop}tuA\selectlanguage{english} & t(\selectlanguage{ethiop}t\selectlanguage{english}) & u(\selectlanguage{ethiop}'u\selectlanguage{english}) & a(\selectlanguage{ethiop}'A\selectlanguage{english}) &  t-u-a \\ [0.05ex]
\selectlanguage{ethiop}^cuA\selectlanguage{english} & \v{c}(\selectlanguage{ethiop}^c\selectlanguage{english}) & u(\selectlanguage{ethiop}'u\selectlanguage{english}) & a(\selectlanguage{ethiop}'A\selectlanguage{english}) &  \v{c}-u-a \\ [0.05ex]
\selectlanguage{ethiop}nuA\selectlanguage{english} & n(\selectlanguage{ethiop}n\selectlanguage{english}) & u(\selectlanguage{ethiop}'u\selectlanguage{english}) & a(\selectlanguage{ethiop}'A\selectlanguage{english}) &  n-u-a \\ [0.05ex]
\selectlanguage{ethiop}~nuA\selectlanguage{english} & \~{n}(\selectlanguage{ethiop}~n\selectlanguage{english}) & u(\selectlanguage{ethiop}'u\selectlanguage{english}) & a(\selectlanguage{ethiop}'A\selectlanguage{english}) &  \~{n}-u-a \\ [0.05ex]
\selectlanguage{ethiop}kuA\selectlanguage{english} & k(\selectlanguage{ethiop}k\selectlanguage{english}) & u(\selectlanguage{ethiop}'u\selectlanguage{english}) & a(\selectlanguage{ethiop}'A\selectlanguage{english}) &  k-u-a \\ [0.05ex]
\selectlanguage{ethiop}zuA\selectlanguage{english} & z(\selectlanguage{ethiop}z\selectlanguage{english}) & u(\selectlanguage{ethiop}'u\selectlanguage{english}) & a(\selectlanguage{ethiop}'A\selectlanguage{english}) &  z-u-a \\ [0.05ex]
\selectlanguage{ethiop}^zuA\selectlanguage{english} & \v{z}(\selectlanguage{ethiop}^z\selectlanguage{english}) & u(\selectlanguage{ethiop}'u\selectlanguage{english}) & a(\selectlanguage{ethiop}'A\selectlanguage{english}) &  \v{z}-u-a \\ [0.05ex]
 . &  & . & & \\ [0.05ex]
. &  & . & & \\ [0.05ex]
. &  & . & & \\ [0.05ex]
\selectlanguage{ethiop}gui\selectlanguage{english} & g(\selectlanguage{ethiop}ge\selectlanguage{english}) & u(\selectlanguage{ethiop}'u\selectlanguage{english}) & i(\selectlanguage{ethiop}'i\selectlanguage{english}) &  g-u-i \\ [0.05ex]
\selectlanguage{ethiop}guE\selectlanguage{english} & g(\selectlanguage{ethiop}ge\selectlanguage{english}) & u(\selectlanguage{ethiop}'u\selectlanguage{english}) & e(\selectlanguage{ethiop}'E\selectlanguage{english}) &  g-u-e \\ [0.05ex]
 \hline

\end{tabular}
\end{center}

\end{document}